\documentclass[10pt,twocolumn,letterpaper]{article}
\pdfoutput=1
\usepackage{wacv}
\usepackage{times}
\usepackage{epsfig}
\usepackage{graphicx}
\usepackage{amsmath}
\usepackage{cuted}
\usepackage{amssymb}
\usepackage{subcaption}
\usepackage{comment}
\usepackage[accsupp]{axessibility}
\def\wacvPaperID{1161} 

\wacvfinalcopy 

\ifwacvfinal
\def\assignedStartPage{9876} 
\fi
\ifwacvfinal
\else
\fi

\ifwacvfinal
\usepackage[breaklinks=true,bookmarks=false]{hyperref}
\else
\usepackage[pagebackref=true,breaklinks=true,colorlinks,bookmarks=false]{hyperref}
\fi

\begin{document}

\title{Robust 3D Garment Digitization from Monocular 2D Images for 3D Virtual Try-On Systems}

\author{Sahib Majithia\\
Myntra Designs Pvt. Ltd.\\
Bangalore, India\\
{\tt\small sahib.majithia@myntra.com}
\and
Sandeep N. Parameswaran\\
Myntra Designs Pvt. Ltd.\\
Bangalore, India\\
{\tt\small sandeep.narayan@myntra.com}

\and
Sadbhavana Babar\\
Myntra Designs Pvt. Ltd.\\
Bangalore, India\\
{\tt\small sadbhavana.babar@myntra.com}
\and
Vikram Garg\\
Myntra Designs Pvt. Ltd.\\
Bangalore, India\\
{\tt\small vikram.garg@myntra.com}
\and
Astitva Srivastava\\
IIIT-Hyderabad,\\
India\\
{\tt\small astitva.srivastava@research.iiit.ac.in}
\and
Avinash Sharma\\
IIIT-Hyderabad,\\
India\\
{\tt\small asharma@iiit.ac.in}
}

\maketitle
\begin{strip}
    \centering
    \includegraphics[width=0.85\linewidth]{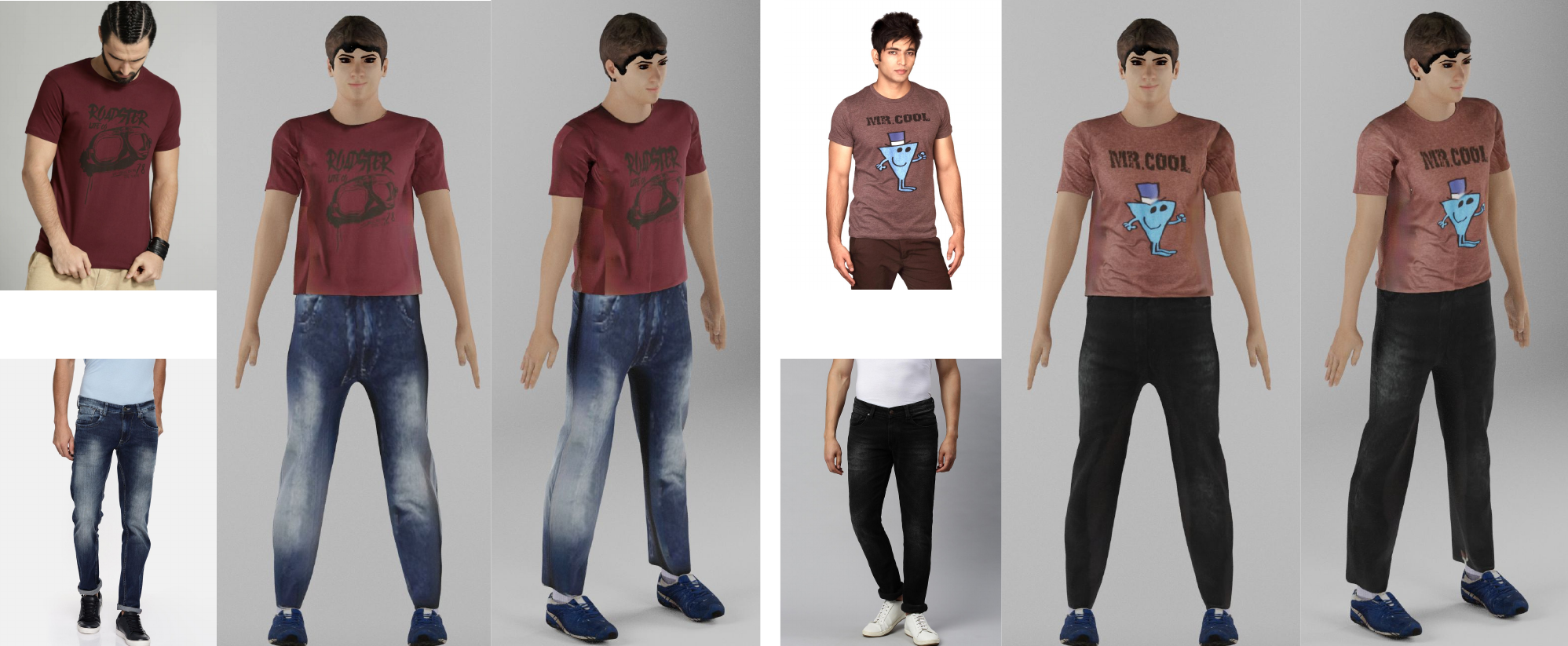}
   \captionof{figure}{Textured template meshes of T-shirts \& Trousers draped on a human avatar.}
\label{fig:teaser}
\end{strip}

\begin{abstract}



%
In this paper, we develop a robust 3D garment digitization solution that can generalize well on real-world fashion catalog images with cloth texture occlusions and large body pose variations. We assumed fixed topology parametric template mesh models for known types of garments (e.g., T-shirts, Trousers) and perform mapping of high-quality texture from an input catalog image to UV map panels corresponding to the parametric mesh model of the garment. We achieve this by first predicting a sparse set of 2D landmarks on the boundary of the garments. Subsequently, we use these landmarks to perform Thin-Plate-Spline-based texture transfer on UV map panels. Subsequently, we employ a deep texture inpainting network to fill the large holes (due to view variations \& self-occlusions) in TPS output to generate consistent UV maps. 
Furthermore, to train the supervised deep networks for landmark prediction \& texture inpainting tasks, we generated a large set of synthetic data with varying texture and lighting imaged from various views with the human present in a wide variety of poses. Additionally, we manually annotated a small set of fashion catalog images crawled from online fashion e-commerce platforms to finetune. We conduct thorough empirical evaluations and show impressive qualitative results of our proposed 3D garment texture solution on fashion catalog images. Such 3D garment digitization helps us solve the challenging task of enabling 3D Virtual Try-on.

\end{abstract}
\section{Introduction}

The fashion e-commerce market is a rapidly growing industry, especially in the post-pandemic era. It is expected to reach 670 billion dollars in 2021 \cite{growth-prediction}. One key challenge faced by this growing industry is the high return rate of around 72\% \cite{high-return-rate}. A large part of these returns are due to customer apprehensions regarding garment ‘look’ and ‘fit' \cite{doi:10.1080/09593969.2017.1314863, 9189849}. A realistic Virtual Try-on (VT) system would help alleviate the high return rate as it will help the customers visualize combinations of different apparel in different sizes on an avatar of their own body shapes. 
However, existing image-based 2D VT solutions \cite{han2018viton, choi2021viton, han2019clothflow, DBLP:journals/corr/abs-2103-09479, xie2021wasvton, issenhuth2019endtoend, wu2019m2etry, ren2021cloth, Minar_CPP_2020_CVPR_Workshops, Jandial_2020_WACV} have inherent limitations in providing a satisfactory user experience as they lack the ability to generate novel poses and view variations for the customer. Majority of 2D VT solutions, which can deal with the variations, use generative models \cite{DBLP:journals/corr/abs-1911-07926, raffiee2020garmentgan, cui2021dressing, yang2020photorealistic} and thus fails to reconstruct images with high-frequency texture details (visual content on clothing) apart from regularly failing on fidelity of exact cloth design features (e.g., neck or collar type). On the other hand, 3D VT solutions use 3D models for clothing and then render a draped version of the same on 3D human avatars (e.g., physics-based cloth deformation by Blender \cite{blender}) of varying body shapes. An important requirement for a 3D VT system is the availability of a large catalog of digitized 3D garment/cloth models, which requires an expensive multi-view capture/scan setup thus can’t scale up for large catalog sizes, especially in fast-fashion scenarios (owing to rapidly changing fashion trends \cite{fastfashion_researchgate, fastfashion_trends}).


Recent advancement in deep learning has enabled attempting this challenging task in generative frameworks. In this line of work, a few recent efforts in the literature \cite{pix2surf,8943678} attempt to automate 3D digitization of clothing from monocular/single catalog image by learning to transfer texture from catalog images to respective 2D texture maps. In  particular,  [22]  learns  a  mapping  from  the input  image  pixels  to  the  UV  map,  in  the  form  of  dense correspondences from garment image silhouettes pixels to the pixels present in the 2D UV map of a 3D garment surface.   Nevertheless,  such mapping is global in nature and does not guarantee to preserve high-frequency textures and tends to blur out fine-grained details like thin lines and curvatures in the cloth texture (as shown in their paper). 
Additionally, \cite{pix2surf} uses pre-segmented ghost mannequin cloth images in the canonical (frontal) pose as target cloth image, whereas a large part of available real catalog images involves actual human models wearing clothing with hands in the different body poses. Another relevant work~\cite{8943678} attempted to retain high-frequency texture details using Thin-Plate-Spline (TPS)~\cite{Berg2006} based texture transfer on cloth UV map panels of template mesh. However, their method can not handle holes and missing parts in texture maps owing to self-occlusions caused by articulated body poses and camera view variations. Their method also requires large-scale annotations of landmarks and segmentation on real data, which is expensive and time-consuming. 

In this paper, we aim to overcome the limitations of existing methods to develop a robust 3D garment digitization solution that can generalize well on real-world catalog images. Such digitization will facilitate a 3D VT system to render a 360-degree view of garments with high-quality textures on realistic custom avatars. 
Our method assumes known types of garments (e.g., T-shirts and Trousers) and uses parametric templates with fixed topology as garment mesh for 3D representation. Thus, the critical task in this setup is mapping high-quality texture from a 2D catalog image to UV map panels of the parametric garment mesh. 
We achieve this by first predicting a sparse set of 2D landmarks on the boundary of the garment and segmentation map of the garment on the 2D image using JPPNet~\cite{liang2018look} architecture trained and finetuned on our synthetic and real dataset. Subsequently, these landmarks are used to perform TPS~\cite{Berg2006} based panel-wise texture transfer on the UV map. We further propose to employ a deep texture inpainting network (MADFNet~\cite{zhu2021image}) to fill in the large holes / missing parts in the UV maps due to view variations and self-occlusions.
Furthermore, a large amount of annotated data is required to train the supervised deep networks for landmark prediction and texture inpainting tasks. However, the majority of the existing fashion data sets (e.g., Deepfashion2~\cite{DeepFashion2}) in this domain are focused on in-the-wild settings instead of the fashion catalog kind of setup. We empirically found that this caused degradation in performance for model trained on such datasets and tested on fashion catalog images. Thus, we generated a large set of synthetic data by varying texture and lighting ($\sim$3000 textures and 6 lighting variations) consisting of mainly T-shirts and trousers garments draped independently on virtual human avatars ($\sim$ 7000 human poses) imaged from various views. Additionally, we also manually annotated a small set ( $\sim$ 1300) of fashion catalog images crawled from online fashion e-commerce platforms and finetuned and test deep networks trained on synthetic data. 
%
We intend to share this synthetic data with the larger academic community. We conducted thorough empirical evaluations and comparisons with SOTA methods and show impressive qualitative results of our proposed 3D garment texture solution on fashion catalog images. 

\section{Literature Review}
\label{sec:literature}
\subsection{2D Techniques}
Jetchev and Bergmann~\cite{jetchev2017conditional} proposed CAGAN, which first introduced the task of swapping fashion articles on human images. VITON~\cite{han2018viton} addressed the same problem by proposing a coarse-to-fine synthesis framework that involves TPS transformation of clothes. Most existing virtual try-on methods tackle different aspects of VITON to synthesize perceptually convincing photo-realistic images. CP-VTON~\cite{wang2018characteristicpreserving} adopted a geometric matching module to learn the parameters of TPS transformation, which improves the accuracy of deformation. VTNFP~\cite{9008110} and ACGPN~\cite{Yang_2020_CVPR} predicted the human-parsing maps of a person wearing the target clothes in advance to guide the try-on image synthesis. CP-VTON+~\cite{Minar_CPP_2020_CVPR_Workshops} improved upon CP-VTON by introducing a concrete loss function at the blending stage to preserve textures, but the authors of CP-VTON+ themselves pointed out that the 2D image-based approach has inherent limitations for coping with diversely posed target human cases. Therefore, the application is limited to simple clothing and standard posed target humans. For diverse cases, 3D reconstruction is more suitable.

\subsection{3D Techniques}

3D human body reconstruction methods, which involve estimating a 3D clothed human body from a monocular RGB image, can be divided into two categories: Parametric and Non-parametric methods. Parametric methods~\cite{zhu2019detailed,ma2020learning,DBLP:journals/corr/abs-2103-06871}  predict deformation from the SMPL~\cite{SMPL:2015} surface to model clothing, but they fail to model high-frequency textures in the clothing. \cite{DBLP:journals/corr/abs-1908-07117} can retain high-frequency texture details but it can not deal with arbitrary topology of loose garments, such as skirts, long dresses etc. Non-parametric methods \cite{saito2019pifu,saito2020pifuhd,huang2020arch,he2020geopifu,DBLP:journals/corr/abs-1904-08645} can generate highly detailed clothed human mesh, but they also tend to produce blurry textures. Moreover, since these methods do not model clothes separately, extracting cloth part from the predicted mesh is not trivial. Additionally, the estimated geometry of the same cloth varies drastically with the pose and shape of the body.

Methods such as \cite{patel20tailornet,gundogdu19garnet,santesteban2021selfsupervised} predict how a synthetic 3D garment dynamically deforms over a given target human body pose and shape. These methods can deal with variations in shape, pose, and style, but they don't emphasize how to accurately transfer textures from real garment images onto the 3D synthetic garment.

Multiple works exist in the literature for creating 3D garment models. Conventional methods of generating 3D garment models use depth cameras \cite{sekine2014virtual,laehner2018deepwrinkles} or multi-view images \cite{bradley2008markerless} which require huge efforts for data capture. \cite{garment_modelling_from_single_image} can recover the underlying 3D mesh of the garment but it doesn't retain texture details, and moreover, it relies on initial human shape and pose estimation which can be unreliable. \cite{physics-inspired-garment-recovery} can also recover 3D mesh which can be re-targeted to other body shapes, but it requires a large database of templates for the human body and garments to estimate parameters for template registration. MGN~\cite{bhatnagar2019mgn} and DeepFashion3D~\cite{zhu2020deep} attempt to reconstruct garment clothing by using real-world 3D scans. \cite{shen2020gan} uses sewing patterns that are edited according to the SMPL body shape given as input. The edited sewing patterns are used to generate 3D garments to be draped on a specific body type. Although this method is robust and requires minimal input, it does not deal with the texture present in the clothing. \cite{pix2surf} allows the transfer of textures to garments by learning to predict dense correspondences using a deep neural network. But, this method also requires images without occlusions and is not generic to every catalog process and does not produce textures at high resolutions.

CLOTH3D~\cite{bertiche2020cloth3d} uses synthetically generated data for the reconstruction of textured 3D meshes. But generalization of models trained on CLOTH3D to real-world images would require 3D scans. DeepGarment~\cite{Dib17a} claims to generalize on real-world samples by training on only synthetic data. But for their method, authors generate 100,000 samples of T-shirts and infer on images of 64x64 resolution. Also, they do not reconstruct textures along with 3D garment shape reconstruction. 

JF-Net~\cite{8943678} automatically transfers textures of clothing images (front and back) to 3D garments of a fixed template by learning a mapping from images to the UV map of the garment. Although, \cite{8943678} attempts to generate textures of 3D garments from RGB images, it restricts itself to images where the garment is completely visible. Pix2Surf~\cite{pix2surf} formulated their solution by using a silhouette mask of the garment. A general e-commerce catalog consists of various brands, which have their specific process for photo-shoot. This generates variations in human poses and occlusions of the garment, which \cite{8943678} cannot handle. For \cite{8943678}  to work, a change in the catalog process is necessary.

\section{Proposed 3D Garment Digitization Solution}
\label{sec:method}

\begin{figure*}
  \includegraphics[width=0.9\textwidth]{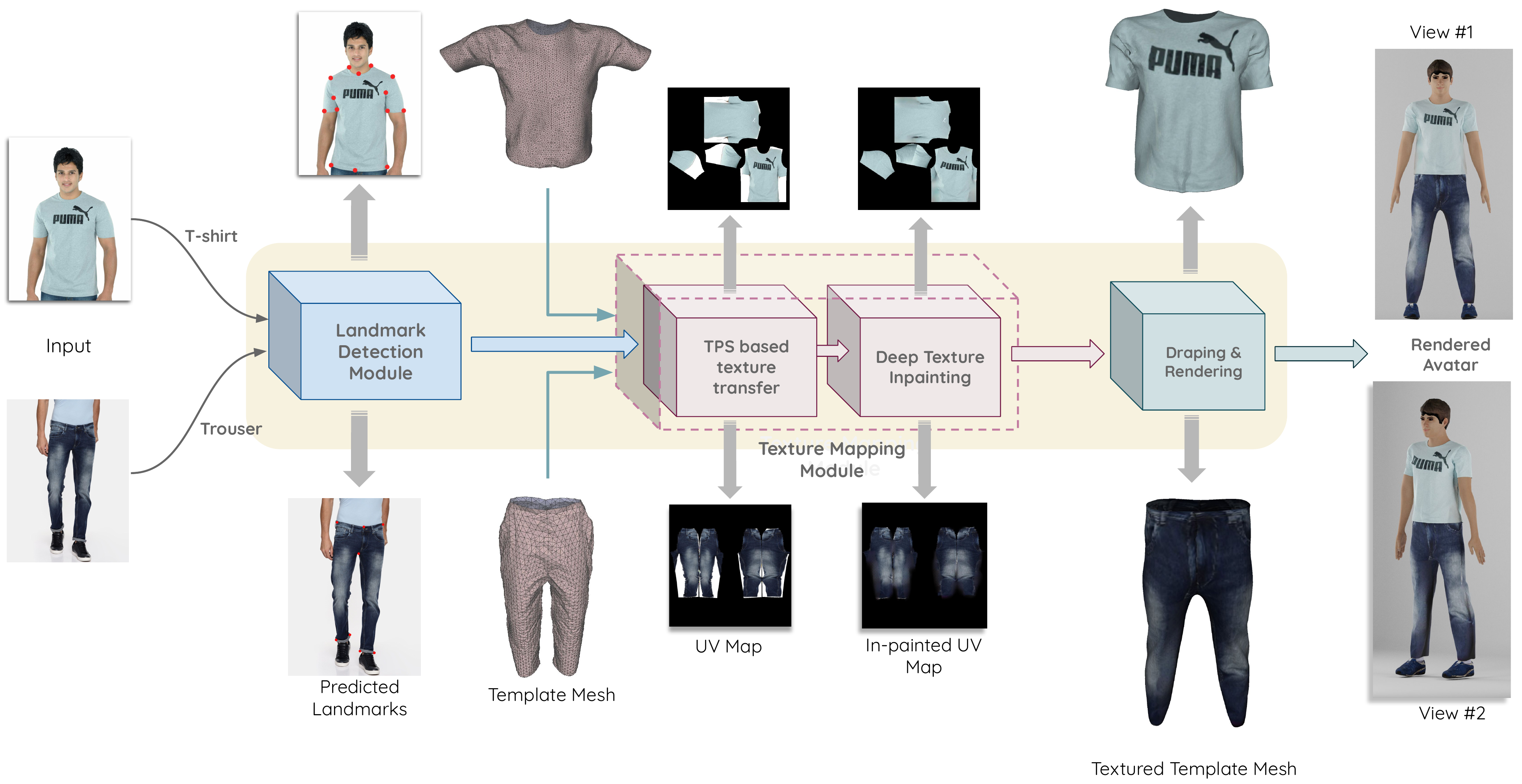}
  \caption{The proposed 3D garment digitization pipeline consists of 1) Landmark Detection Module which predicts a set of 2D landmarks and associated garment segmentation map, 2) Texture Mapping Module uses the predicted landmarks to transfer the texture onto the UV map of the template mesh using Thin-plate-Splines followed by inpainting on the self-occluded regions of the UV map using the deep network and 3) Draping \& Rendering Module drapes the template mesh along with inpainted UV map on the target 3D human avatar. (Here we show results on T-shirts and Trousers simultaneously, but the whole process is carried out for each garment type separately. In the end, both garments are draped together.)}
  \label{fig:architecture}
\end{figure*}

Here, we introduce our novel solution for 3D garment digitization from generic e-commerce catalog images having body pose variations causing self-occlusions to garments. 
We assumed fixed topology parametric template mesh models for known types of garments (e.g., T-shirts) and perform mapping of high-quality texture from an input catalog image to UV map panels corresponding to the parametric mesh model of the garment.  The architecture diagram is as shown in Figure \ref{fig:architecture}. We achieve this by first predicting a sparse set of 2D landmarks/key points on the boundary of garments in the image. We adapt the existing JPPNet~\cite{liang2018look} architecture, which was originally proposed for human body joint prediction, for garment landmark prediction. Subsequently, these landmarks are used to perform TPS~\cite{Berg2006} based texture transfer on UV map panels. However, due to view variations and self-occlusions (as discussed before), these UV maps have large holes/missing parts. Additionally, clothing semantic segmentation maps predicted by JPPNet are used to remove pixels within the landmark boundary that are labeled as background or fashion articles occluding the clothing. Hence, we propose to employ a deep texture inpainting network (MADFNet~\cite{zhu2021image}) to fill the large holes / missing parts to generate consistent UV maps. This network takes as input each panel in the UV along with the occluded regions and inpaints relevant texture using the information available from the non-occluded parts of the image. 

\begin{figure}
  \includegraphics[width=\linewidth]{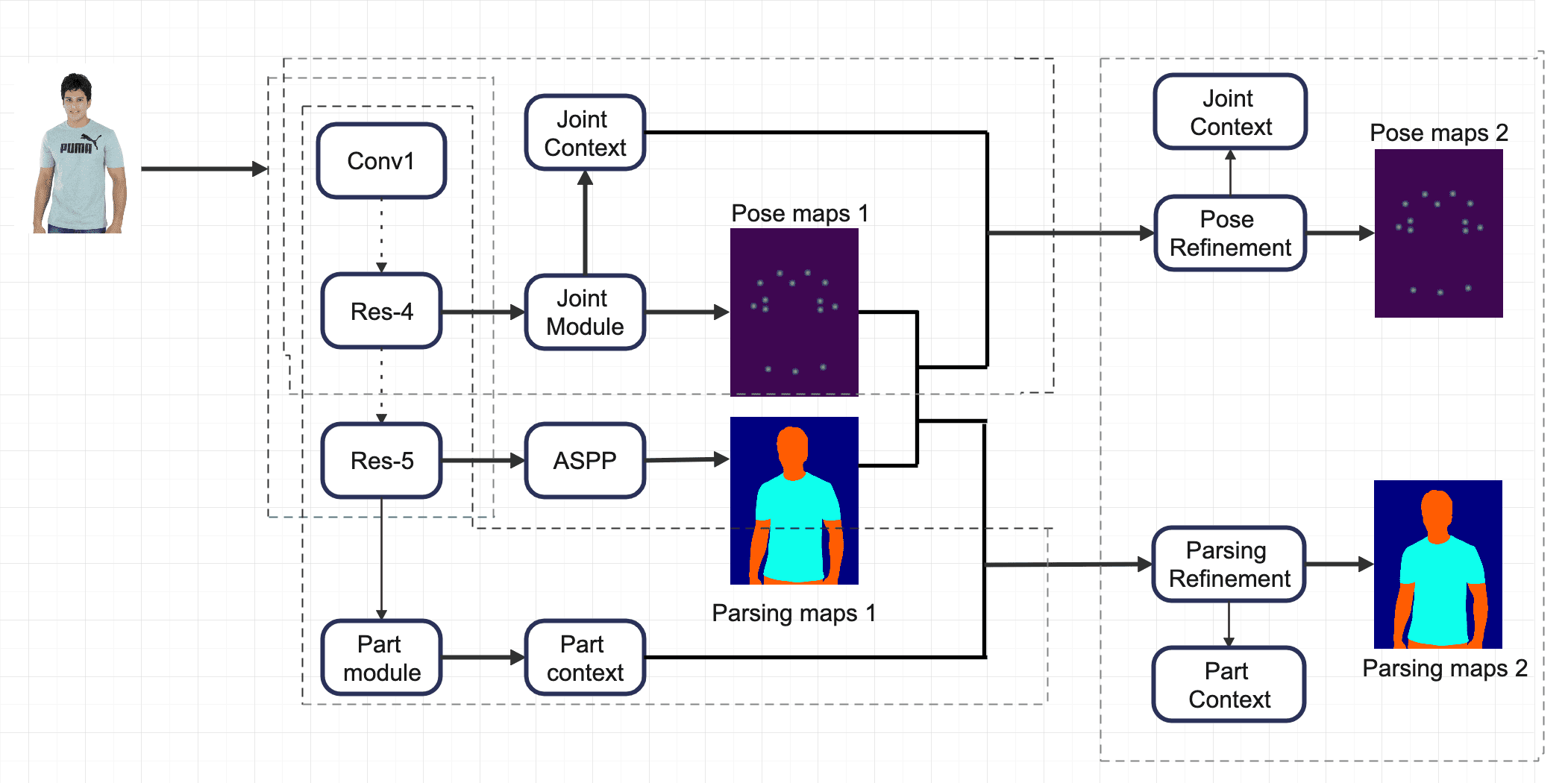}
  \caption{The architecture of JPPNet~\cite{liang2018look} adapted in our pipeline}
  \label{fig:JPPMADF-Arch}
\end{figure}

\subsection{Landmark Detection Module}
\label{sec:lpnet}
We adapt the original JPPNet~\cite{liang2018look} architecture, as shown in Figure \ref{fig:JPPMADF-Arch}, to jointly predict landmarks and semantic maps. The landmarks for each garment are defined following JF-Net~\cite{8943678}. The number of landmarks varies with garment type (e.g. 14 for the T-shirts and 8 for the Trousers). Their original JPPNet architecture was proposed for the human parsing and pose estimation, we adopt it for our task of clothing landmark prediction. We use clothing segmentation similar to the original paper.  The architecture uses a common encoder and separate branches for both tasks, features of which are concatenated in the refinement  stage to  utilize  the  coherent  representation of cloth parsing and landmark prediction.  Similarly, in our multi-task learning setup, the parse maps produced by the network guide the predictions of landmarks as both tasks are spatially aligned. 
In addition to this, the network uses an iterative refinement scheme to aid individual  tasks. 
Although we  use  both  tasks  for  further stages, the texture mapping module is influenced more by pixel-level alignment of the landmarks in comparison to the segmentation maps. Thus, we focus our attention and evaluations on the landmark prediction task.
%

\subsection{Texture Mapping Module}
\label{sec:inpaint}


This module takes predicted landmarks as input and maps the relevant regions of the 2D image onto the UV map of the template garment mesh for texture transfer. Particularly, we use the inferred landmarks as the control points for the Thin-Plate-Spline~\cite{Berg2006} transformation. Each landmark in the 2D image is also mapped to a fixed point in UV map of the template garment mesh, which is mapped manually, once for each template based on the structure of panels that compose the template. The region of the 2D image to be used for texture transfer is masked using the segmentation maps provided by the Landmark Detection Module.
In comparison to the generative deep learning approaches that produce blurry textures, TPS enables us to transfer high-frequency texture details and provides a pixel-level accurate mask required for texture inpainting. The output of TPS yields partially filled UV maps with large missing parts/holes owing to self-occlusion and view-variations. In order to recover consistent UV maps, we employ automated texture inpainting network MADFNet~\cite{zhu2021image}. It is an encoder-decoder-based architecture that comprises a Mask-Aware Dynamic Filtering (MADF) Module that aids in filling arbitrary missing regions. This network was originally proposed for in-the-wild images and trained on the Places dataset~\cite{zhou2017places}. We propose to finetune the model (separately for each garment type) for the UV maps of the synthetically generated data.
The training and test data for the inpainting network consist of multiple panels for each type of clothing. 

Generation of back panels of T-shirts is severely ill-posed the back camera is missing. Thus, we can either take into account the front panel and copy it to the back or take a uniform gradient patch from the front panel and perform texture copying to the back panel by replicating that patch on the entire back panel. However, if the back view image is readily available (as a part of the catalog), we can generate the back panel by using landmarks predicted on the back view of the T-shirt. In our experiments, we used the first strategy for filling back panel texture.

\subsection{Draping and Rendering Module}
\label{subsec:draping}
Given the fixed template for the T-shirt and trousers alongside a human avatar, each garment is aligned with the avatar in T-pose independently. A cloth deformation is obtained for the garments by applying a cloth simulation~\cite{bertiche2020cloth3d, patel20tailornet} to the template garment using Blender~\cite{blender}. 
We apply a motion sequence to animate the human avatar and deform the garment corresponding to a target pose. The motion sequence has three keyframes: the first one where the avatar is in T-pose, an intermediate keyframe where the avatar remains in T-pose, and a final keyframe where the avatar is reposed to the target pose (we use an A-pose for the final 3D VT) using Linear Blend Skinning (LBS). A simple LBS skinning was used since the poses involved in the e-commerce cataloging process were largely devoid of extreme poses, and hence adequate realism was maintained.
The human pose is kept constant for the initial set of frames to allow the cloth deformation to stabilize. The reposing from T-pose to the target pose is also done in multiple frames for enabling the cloth simulation to walk over multiple intermediate steps to obtain a realistic garment deformation. We individually apply the deformations to each template and perform a collision resolution to visualize both garments together on a human avatar.

\begin{figure}[t]
\begin{center}
   \includegraphics[width=\linewidth]{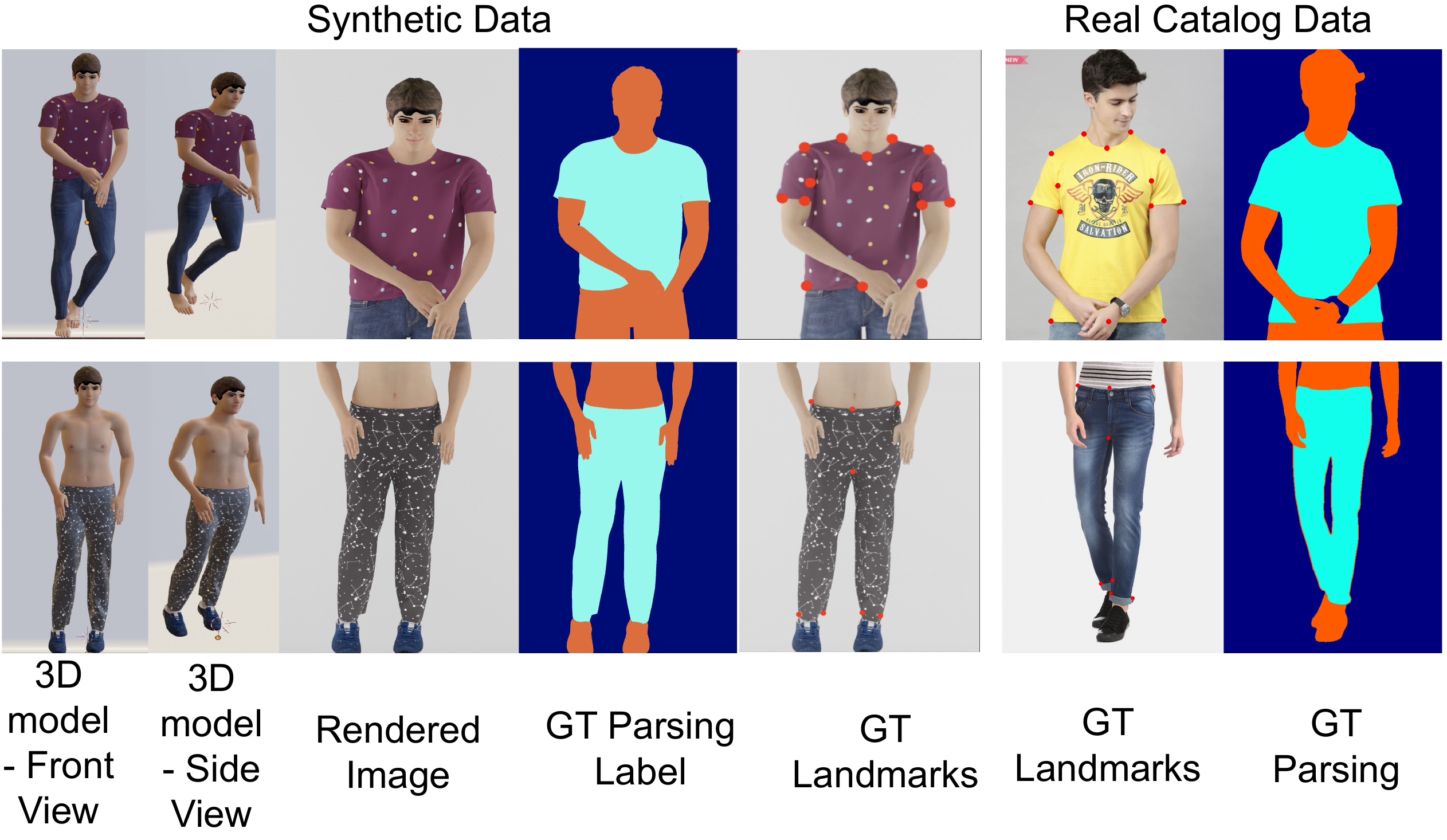}
\end{center}
   

   \caption{\textbf{Synthetic Data} and \textbf{Real Catalog Data} samples along with GT annotations.}
\label{fig:overview}
\end{figure}

\begin{figure}[t]
\begin{center}
   \includegraphics[width=\linewidth]{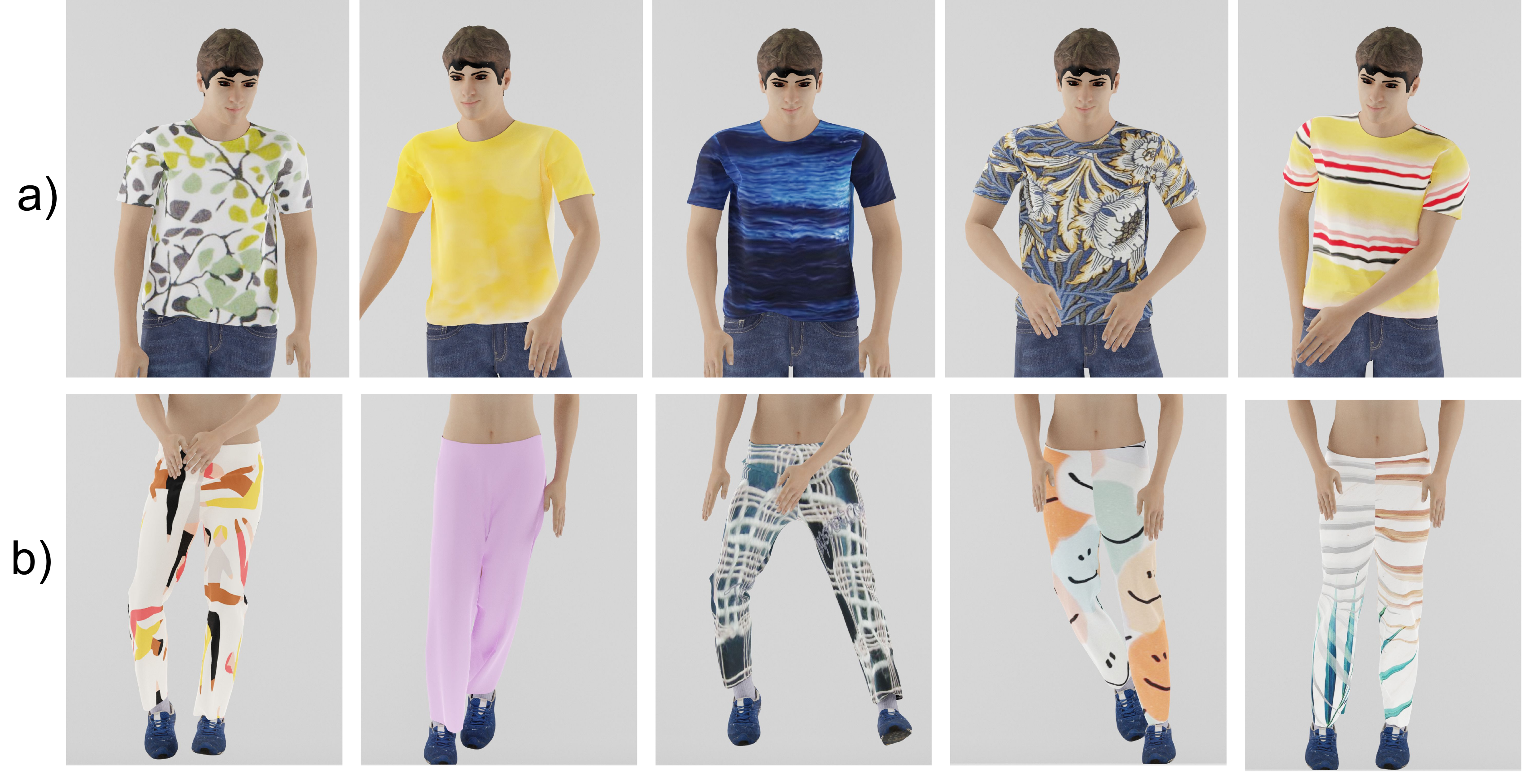}
\end{center}
   \caption{Sample images from our synthetic dataset highlighting diverse textures and poses for (a) T-shirts and (b) Trousers}
   
\label{fig:long}
\label{fig:dataset_variance}
\end{figure}
\section{Experiments \& Results}
\label{sec:experiments}

\subsection{Our Synthetic Dataset}
\label{sec:dataset}


We generated synthetic data and rendered the images in a way that resembles an e-commerce catalog. We have taken template 3D garment meshes from the Berkley Garment Library~\cite{de2012berkeley} and human avatar meshes created using MakeHuman~\cite{makehuman} that are rigged with a CMU MB armature. We utilized a T-shirt template and a trousers template for our use case. We included variations in the textures applied to the template meshes, poses of the human avatar, lighting of the scenes and minor variations in template size and camera coordinates. Over 3000 textures in the form of images were crawled from the internet to be applied to the template meshes. We also crawled over 7000 e-commerce catalog full shot images (i.e images that have the full human body visible) to generate a wide range of human poses. We further estimated the 3D human pose of each of these crawled images using \cite{li2020cascaded}. The estimated skeleton information from the 3D pose was converted to joint angles and written in MoCap format~\cite{cmu_mocap}. The human avatar was further re-targeted to the estimated 3D pose by generating the rotation and translation parameters for each bone present in the armature of the rigged skeleton with respect to the saved skeleton present in the form of motion data. The target poses were restricted to a near frontal pose by limiting the rotation matrix of the ``Hips'' and ``Head'' bones present in the CMU MB armature. The garments were draped as described in Sec.\ref{subsec:draping} with the target poses obtained from estimated poses from the full-shot images (see Figure \ref{fig:overview}). The synthetic images were rendered using Blender CYCLES in 1080 $\times$ 1440 resolution. We also included 6 lighting conditions for rendering the scene by placing the multiple lights at different positions in the scene, mimicking an e-commerce cataloging activity. Few samples from the our synthetic dataset are shown in Figure \ref{fig:dataset_variance}. 

Each generated image was paired with segmentation maps and landmark annotations. These annotations were obtained similarly as in \cite{cartucho2020visionblender}. The T-shirt template mesh was associated with 14 landmark vertices and the trousers template mesh with 8 vertices. The corresponding 2D landmark annotations of these vertices in the rendered images were obtained by projecting the vertices onto the image using camera parameters. The segmentation labels were also obtained by labeling the garment in different rendering passes.

\begin{figure}[th]
  \includegraphics[scale=0.35]{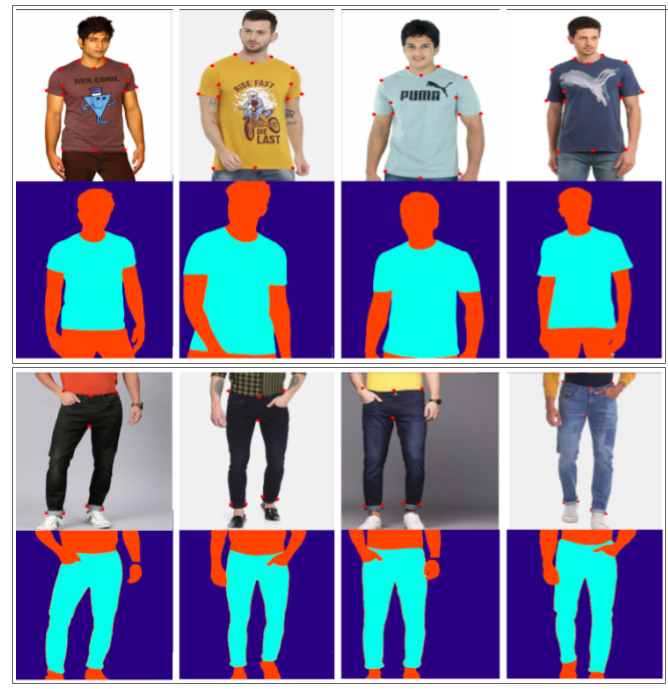}
  \caption{Landmarks and segmentation maps predicted by the landmark detection module on real catalog images.}
  \label{fig:landmark-qualitative}
\end{figure}
%
\begin{figure}[th]
  \includegraphics[width=\linewidth]{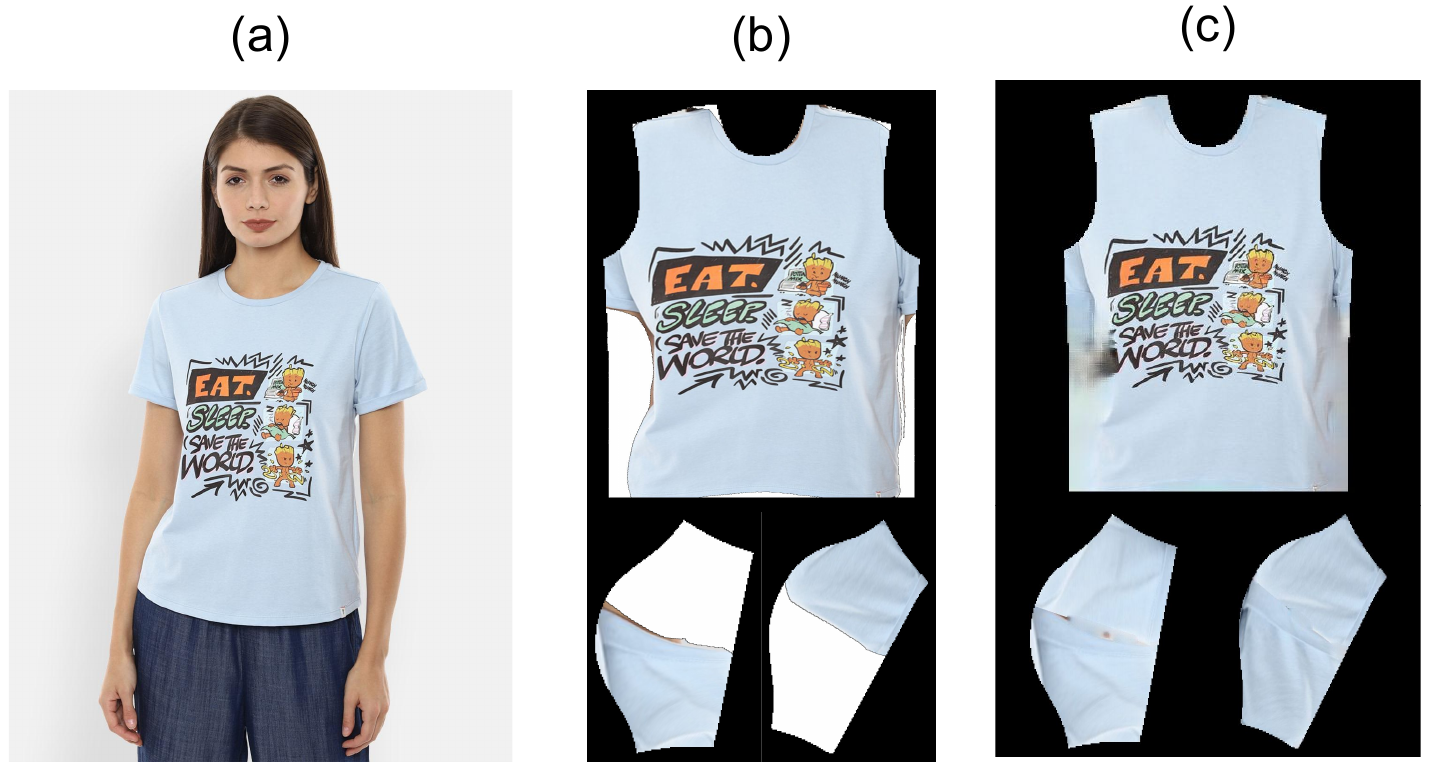}
  \caption{\textbf{Texture mapping module:} \textbf{(a)} Input image, \textbf{(b)} Texture transfer through TPS for the front panel and sleeves, \& \textbf{(c)} Inpainted panels.}
  \label{fig:texture_mapping}
\end{figure}

\subsection{Our Real-World Catalog Dataset}
%
We manually annotated the landmarks on real-world catalog images. We crawled 1300 images for each type of garment from the web, out of which 1000 images were used for training and 300 for validation.
Since the manually annotated landmarks are subjective in nature and prone to human mistakes, we also estimated the errors produced by the manual annotators. For this purpose, 200 images were sampled from the data and annotated by 5 different annotators and the estimated error values are compared with the prediction errors of the landmark detection module. The clothing segmentation labels are generated using \cite{liang2018look}.
\begin{figure*}[thb]
\begin{center}
   \includegraphics[width=0.9\linewidth]{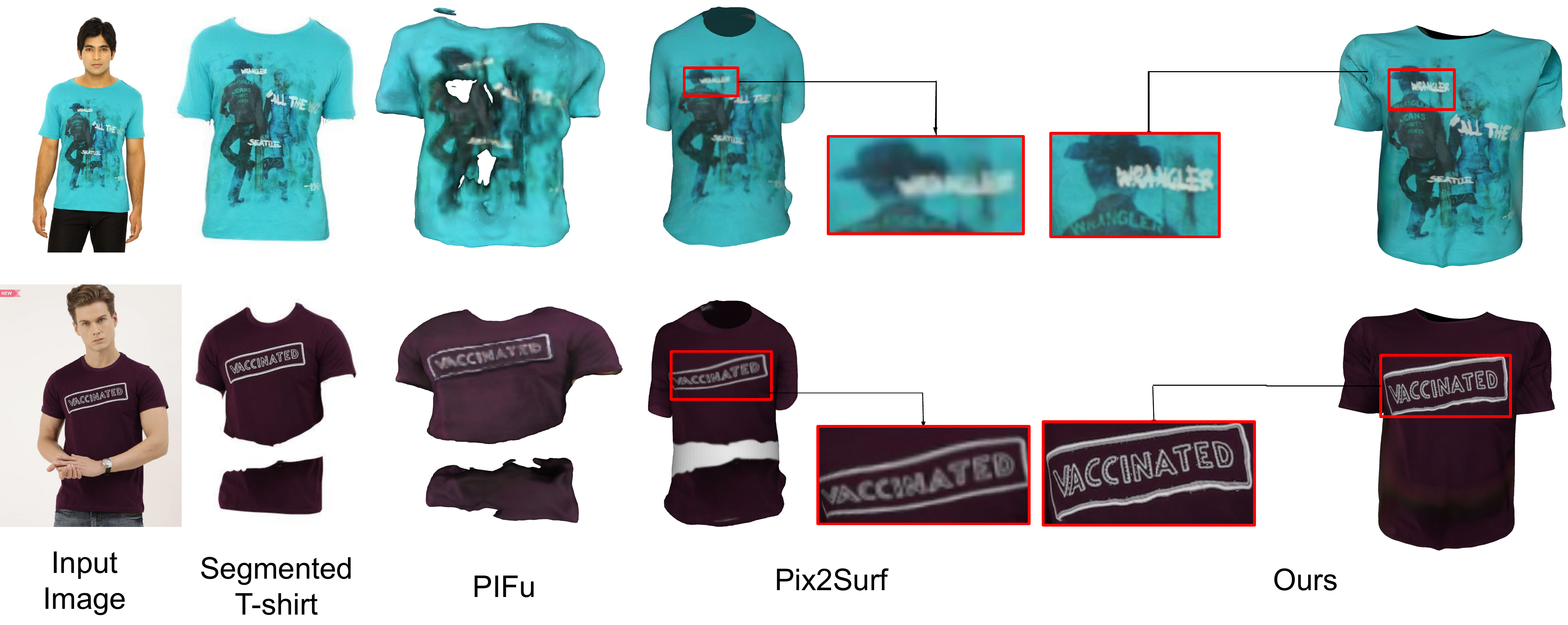}
\end{center}
   \caption{Qualitative comparison of our approach with Pix2Surf~\cite{pix2surf} and PIFu~\cite{saito2019pifu}.}
\label{fig:comparison}
\label{fig:comparison_overview}
\end{figure*}

\subsection{Training Protocol}
\label{sec:training}
%
We train separate networks for T-shirts and Trousers clothing. 
For the landmark detection task, we adapted the original JPPNet architecture and train the network with 384x384 resolution images in two settings, namely, {\bf scratch} and {\bf finetune} with varying sizes of training data. The former refers to training the model on our real dataset starting from random initial weights. The latter refers to training with synthetic data first and further finetuning of the network parameters using images from our Real-World Catalog Dataset. 

In regard to texture inpainting network, we finetune the original MADF-Net~\cite{zhu2021image} on our synthetic data. Our dataset has 3000 images and 7300 masks for T-shirt-panels and 12000 images and 12000 masks for trousers panels. We have trained the network for both T-shirts and trousers panels for 20 epochs. Since, the back part of the sleeves is occluded in the frontal pose, for inpainting the sleeve panel we assumed that sleeves are symmetric. 
\subsection{Results}
\label{sec:results}
%
First, we discuss the results of our landmark detection module. 
We use normalized mean squared error (NMSE)~\cite{liu2016fashion} to evaluate our Landmark detection network and report quantitative results in Table~\ref{tab:landmark-eval}. We compare \textbf{JPPNet(scratch)-1000} and \textbf{JPPNet(synthetic+finetune) -1000} on the same test set of 300 images (1000 denotes number of \textbf{real images} used during training/finetuning). Since the authors of JFNet~\cite{8943678} did not provide code and dataset for evaluation, we report the results from their paper. However, the comparisons are made for the same set of landmarks for comparison. The performance of our JPPNet (scratch)-1000 is comparable to JFNet* -2000 (uses 2000 images for training), although we use a significantly lesser number of real-world training images (almost 50\% less). We further see that a network trained on synthetic data and finetuned on real-world data performs almost twice as good as JFNet in terms of NMSE. %
Another important observation to note is that the errors in our best performing model i.e., JPPNet (synthetic+finetune)-1000 are of the same order as that of human annotations, which have an NMSE of 0.006 primarily contributed by the difference in the exact position of the landmarks among different annotators.
We also conducted an ablation study by varying the size of training data and reported NMSE values, as shown in Table~\ref{tab:landmark-ablation}. We trained with 250, 500 and 1000 real-world training images in scratch and finetuned settings, e.g., JPPNet (synthetic + finetune)-250 uses 5000 synthetic images for initial training and 250 real images for finetuning. We can observe that JPPNet (synthetic + finetune)-250, outperforms JPPNet(scratch) -1000 thereby establishing the fact that training on synthetic data significantly improves the performances while reducing the dependence on expensive manually annotated real-world data. 
Figure~\ref{fig:landmark-qualitative} shows the qualitative results on landmark detection and segmentation task on real-world test images. As we can observe, our model predicts accurate landmark locations and smooth and precise segmentation maps.

\begin{figure}
  \includegraphics[width=0.9\linewidth]{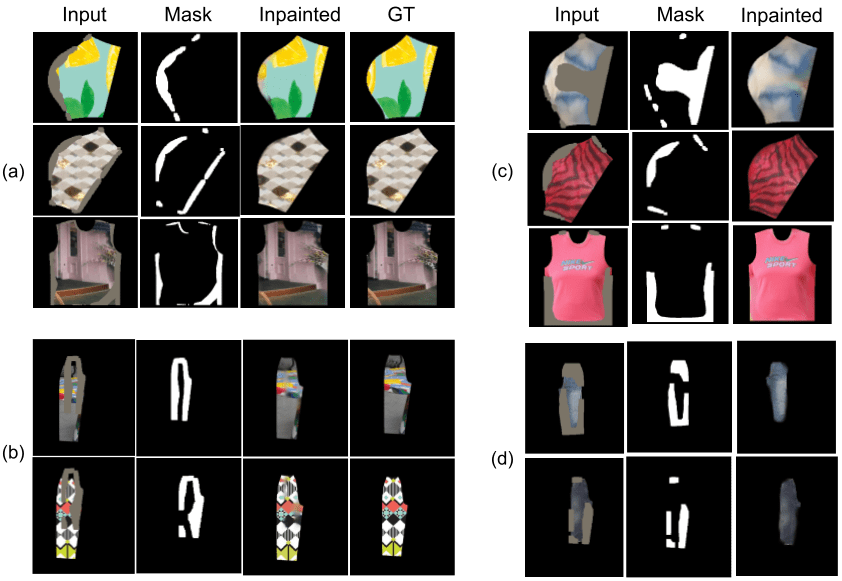}
  \caption{\textbf{Qualitative results of inpainting network:} on synthetic data \textbf{(a)} \& \textbf{(b)} and on real-world data \textbf{(c)} \& \textbf{(d)} (GT is not available for real data.)}
\label{fig:inpainting}
\end{figure}

Further, we discuss the qualitative and quantitative results of our texture inpainting network.
We use widely used image similarity metrics PSNR and SSIM (as used in~\cite{hore2010image}) for quantitative evaluation.  Table~\ref{tab:inpaint-syn} shows the quantitative evaluation of networks finetuned on our synthetic data in comparison to the original network (pretrained on the Places2~\cite{zhou2017places} dataset) inferred on our synthetic data. We can observe that finetuning on our synthetic data significantly improves the reported performance on both the metrics on our test data (981 sample images for T-shirts and 1187 sample images for trousers). Figure~\ref{fig:texture_mapping} shows the output of TPS-based texture transfer followed by the output of the texture inpainting network.
Figure \ref{fig:inpainting} (a) and (b) shows the result of the inpainting network trained on our synthetic data.  We also test our finetuned models on real data and portray the generalizability of our models trained on synthetic data as depicted in Figure \ref{fig:inpainting} (c) and (d). 

\begin{table}
\begin{center}
\begin{tabular}{|l|c|c|}
\hline
 Method & T-shirts &  Trousers \\
\hline\hline

JFNet* -2000 & 0.031 & 0.022 \\ 
\hline
JPPNet (scratch) -1000 & 0.029 & 0.026 \\
JPPNet (synthetic+finetune) -1000 & 0.014 & 0.012 \\

\hline
\end{tabular}
\end{center}
\caption{Evaluation of the landmark detection module using normalized mean squared error (lower the better). * denotes results reported using JFNet evaluated on a different dataset.}
\label{tab:landmark-eval}
\end{table}

\begin{table}
\begin{center}
\begin{tabular}{|l|c|c|}
\hline
 Method variant & T-shirts &  Trousers \\
\hline\hline
scratch -250 & 0.192 & 0.212 \\
scratch -500 & 0.095 & 0.102 \\
scratch -1000 & 0.029 & 0.026 \\
finetune -250 & 0.022 & 0.018 \\ 
finetune -500 & 0.015 & 0.014 \\
finetune -1000 & 0.014 & 0.012 \\
\hline
\end{tabular}
\end{center}
\caption{Performance of  landmark detection module on different variants of the dataset used for training. }
\label{tab:landmark-ablation}
\end{table}

\begin{table}
\begin{center}
\begin{tabular}{|l|c|c|c|}
\hline
 Method & Clothing & PSNR &  SSIM \\
\hline\hline

MADF-Net pretrained & T-shirts & 41.16  & 0.946 \\ 
MADF-Net finetuned & T-shirts & \textbf{42.41} & \textbf{0.977} \\ 

MADF-Net pretrained & Trousers & 46.22 &  0.985\\ 
MADF-Net finetuned & Trousers & \textbf{46.39} &  \textbf{0.99}\\ 
\hline
\end{tabular}
\end{center}
\caption{Quantitative results of texture inpainting on synthetic data (Higher the better for both PSNR and SSIM).}
\label{tab:inpaint-syn}
\end{table}





Since there is no ground-truth data for real world catalog images, and the garment mesh geometry varies across methods, we perform a qualitative comparison following~\cite{pix2surf}. We qualitatively compare our method with two existing approaches PIFu~\cite{saito2019pifu} and Pix2Surf~\cite{pix2surf} (which also predicts the UV map for the garment meshes) in Figure \ref{fig:comparison_overview}. As can be seen clearly, PIFu~\cite{saito2019pifu} and  Pix2Surf~\cite{pix2surf} failed to preserve the high-frequency texture details, whereas our method can preserve such fine texture details. Additionally, PIFu lacks the ability to model the geometry of the cloth separately. The bottom row in Figure \ref{fig:comparison_overview} shows that unlike PIFu and Pix2Surf, our method can handle self-occlusions.

\section{Discussion}
\label{sec:discussion}
We have shown the effectiveness of our method on both T-shirts and Trousers. However, it will be interesting to explore an extension of our approach on loose clothing scenarios (e.g., skirts and gowns) as well as relaxing the template mesh constraint for modeling free-form clothing and garment sizing. 

%
\begin{figure}[h]
  \includegraphics[width=\linewidth]{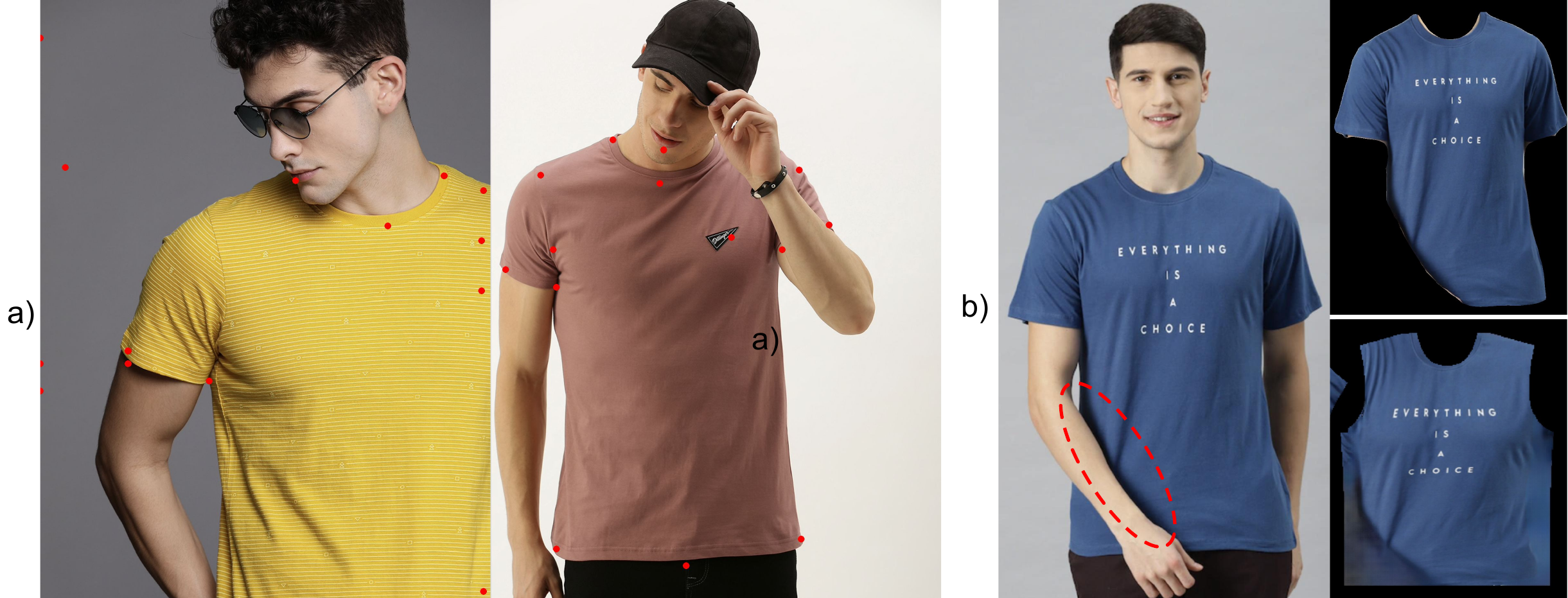}

  \caption{\textbf{Failure Cases: } of Landmark Detection Module  \textbf{(a)}, and  Texture Mapping \& Inpainting Module \textbf{(b)}.}

\label{discussion_failure_cases}
\end{figure}
Our landmark detection module can achieve good performance on the NMSE metric, where we observed that the error reported by our method is close to that of a human annotator. However, our method doesn't perform well on those real catalog images which are significantly far from the training data distribution of synthetic images, for e.g.if the region of interest is cropped out or severe self-occlusion is present, as shown in Figure \ref{discussion_failure_cases} (a). 
The deep inpainting network, which is trained only on synthetic data, can handle hand occlusions well and can also fill in the missing information along the border of individual panels in the UV map. Although a standard cataloging procedure tries to avoid occluding the regions of the garment with fine details, real world catalog images can have occlusions, along with extreme poses and shadows for enhanced realism. In such non-standard scenarios, the inpainting network fails to achieve the level of perfection needed. One such instance is shown in Figure \ref{discussion_failure_cases} (b). The presence of wrinkles and folds in the real-world catalog images in comparison to our synthetic dataset aggravates the challenge in such scenarios. 
In regard to normalization for the garment sizes, since we have employed a fixed template garment mesh, the final digitized garment is robust to further modelling of geometrical deformations due to changes in human pose, body shape and garment sizes, that are essential for applications such as 3D VT. Our proposed TPS based texture transfer might yield slightly inaccurate texture mapping in regions with complicated geometrical structure or perspective distortions. Nevertheless, given the pose variations of fashion catalog images, such occurrences will be minimal.








%
\section{Conclusion \& Future Work}
\label{sec:conclusion}
We propose a deep learning-based 3D garment digitization technique that is robust and can generalize well to real-world fashion catalog images. 
We developed a data generation pipeline that is capable of generating synthetic data with diverse poses and textures for any garment style. This synthetic data helped us train a landmark detection network and a texture inpainting network that is then finetuned with a relatively smaller number of real annotated images. We observed that our training strategy allows both of the networks to generalize well to unseen real images, indicating that our synthetic data is diverse enough to capture the characteristics of real-world images. The proposed pipeline is compact and scalable. The template-based digitization of the garment allows the transfer of the texture easily and avoids loss in texture quality. As part of future work, it will be interesting to explore domain adaptation techniques to further exploit learning from synthetic data. 

{\small
\bibliographystyle{ieee_fullname}
\bibliography{egbib}
}

\end{document}